\documentclass{article}

\usepackage{graphicx}
\usepackage{subfigure}
\usepackage[preprint]{neurips_2023}

\usepackage[utf8]{inputenc} % allow utf-8 input
\usepackage[T1]{fontenc}    % use 8-bit T1 fonts
\usepackage{hyperref}       % hyperlinks
\usepackage{url}            % simple URL typesetting
\usepackage{booktabs}       % professional-quality tables
\usepackage{amsfonts}       % blackboard math symbols
\usepackage{nicefrac}       % compact symbols for 1/2, etc.
\usepackage{microtype}      % microtypography
\usepackage{xcolor}         % colors

\title{GoalieNet: A Multi-Stage Network for Joint Goalie, Equipment, and Net Pose Estimation in Ice Hockey \textnormal{(Extended Abstract)}}

\author{%
  Marjan Shahi\\
  Department of Systems Design Engineering\\
  University of Waterloo \\
  Waterloo, ON N2L3G1 \\
  \texttt{marjan.shahi@uwaterloo.ca} \\
  \And
  David Clausi \\
  Department of Systems Design Engineering \\
  University of Waterloo, ON, CAN \\
  \texttt{dclausi@uwaterloo.ca} \\
  \AND
  Alexander Wong\\
  Systems Design Engineering\\
  University of Waterloo, ON, CAN\\
  \texttt{alexander.wong@uwaterloo.ca} \\
}

\begin{document}

\maketitle

\begin{abstract}
In the field of computer vision-driven ice hockey analytics, one of the most challenging and least studied tasks is goalie pose estimation. Unlike general human pose estimation, goalie pose estimation is much more complex as it involves not only the detection of keypoints corresponding to the joints of the goalie concealed under thick padding and mask, but also a large number of non-human keypoints corresponding to the large leg pads and gloves worn, the stick, as well as the hockey net. To tackle this challenge, we introduce GoalieNet, a multi-stage deep neural network for jointly estimating the pose of the goalie, their equipment, and the net. Experimental results using NHL benchmark data demonstrate that the proposed GoalieNet can achieve an average of 84\% accuracy across all keypoints, where 22 out of 29 keypoints are detected with more than 80\% accuracy. This indicates that such a joint pose estimation approach can be a promising research direction. 
\end{abstract}

\section{Introduction}
\label{sec:intro}

The task of goalie pose estimation is underrepresented from a research perspective in the field of ice hockey sports analytics. Pose estimation is essential to investigate for ice hockey analytics because interpreting pose is key to making action recognition possible which is subsequently necessary for automated player evaluation. Due to the special shape of the gear and equipment the goalie uses, as well as the need for finding the pose of the net, the typical human keypoints found in widely used datasets such as COCO~\cite{lin2014microsoft} are not practical. More specifically, the joints of a goalie are hidden under heavy padding, making conventional human pose estimation training and datasets poorly suited for this task. Furthermore, new keypoints are needed based on the shape of the mask, pads, and gloves, as they are very different from typical human keypoints.

\begin{figure}[htbp]
\centering
\includegraphics[width=\textwidth]{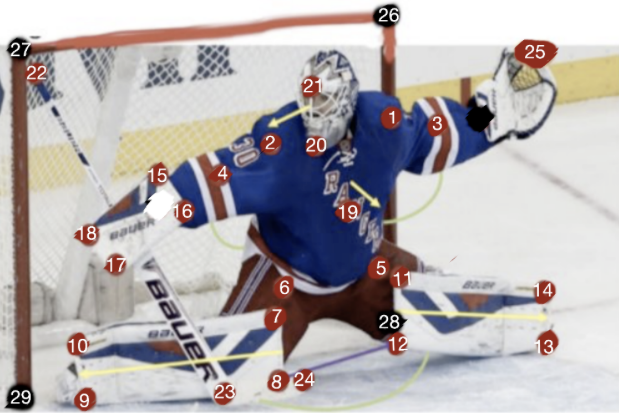}
\caption{The picture illustrates the position of each keypoint for the goalie, equipment, and net. The assigned number and name of each keypoint are presented in Table~\ref{tab:joints}}
\label{fig:goalie}
\end{figure}

Moreover, the need to track the net, which can move during a game, also means the need to detect non-human keypoints corresponding to the shape of a net. All these factors make goalie pose estimation very different from what is generally studied in the common approaches of human pose estimation~\cite{Zhang_2019_CVPR, 7961778}, and require a much larger mix of human and non-human keypoints to cover the goalie, equipment, and net (see Fig.~\ref{fig:goalie}). 

To address these problems, we propose GoalieNet, a multi-stage network for jointly estimating the pose of a goalie along with their stick and the net in a simultaneous manner.

\section{Related Works}
\label{sec:lit}

Pose estimation is a common application for machine learning. To benefit the localization of keypoints, Xu et al.~\cite{xu2022adaptive} use the semantic relationship between different joints. The importance of capturing the relationship between features has been considered by Wang et al.~\cite{wang2020graph} as well. To account for occlusion, learning to predict poses via tracklets is the proposed approach by Yang et al.~\cite{yang2021learning}. The effect of various design factors on the performance of multi-stage approaches is investigated by Li et al.~\cite{li2019rethinking}. Research literature related to pose estimation in ice hockey is scarce. Player keypoints are estimated through a stacked hourglass network by Neher et al.~\cite{neher2017pose}. Building on this, HyperStackNet by Neher et al.~\cite{8575769} jointly estimates player and stick poses. 

\section{Methodology}
\label{sec:method}

In this section, we will first describe our dataset and the different keypoints, followed by a description of the proposed GoalieNet network. 

\subsection{Dataset}
The dataset used comprises 32 NHL video clips, with 5032 frames containing a goalie. The keypoints being detected are listed in Table~\ref{tab:joints} and shown visually in Fig.~\ref{fig:goalie}. In addition to 22 keypoints related to the goalie, are four keypoints for the net and three keypoints for the stick, making our data different from the common dataset for human pose estimation.

\renewcommand{\arraystretch}{1.3}
\begin{table}[htbp]
    \caption{The keypoints for the goalie, equipment, and net are assigned specific numbers and names which are presented in this table. The Fig.~\ref{fig:goalie} depicts the position of each keypoint along with its corresponding number.}
    \vspace{10pt}
    \centering
    \label{tab:joints}
    \begin{tabular}{|p{0.25cm}|l|p{0.25cm}|l|p{0.25cm}|l|}
      \hline
      1 & Left Shoulder & 11 & Left Legpad(0) & 21 & Mask-High \\
      \hline
      2 & Right Shoulder & 12 & Left Legpad(1) & 22 & Stick-Upper\\
      \hline
      3 & Left Elbow & 13 & Left Legpad(2) & 23 & Stick-Lower\\
      \hline
      4 & Right Elbow & 14 & Left Legpad(3) & 24 & Stick-Blade-Tip\\
      \hline
      5 & Left Hip & 15 & Blocker(0) & 25 & Mit-Top\\
      \hline
      6 & Right Hip & 16 & Blocker(1) & 26 & Net-Top-Left\\
      \hline
      7 & Right Legpad(0) & 17 & Blocker(2) & 27 & Net-Top-Right\\
      \hline
      8 & Right Legpad(1) & 18 & Blocker(3) & 28 & Net-Bottom-Left\\
      \hline
      9 & Right Legpad(2) & 19 & Torso Center & 29 & Net-Bottom-Right\\
      \hline
      10 & Right Legpad(3) & 20 & Mask-Low &  & \\
      \hline
    \end{tabular}
    \label{tab:my_label}
\end{table}

\begin{figure}[htbp]
\centering
\includegraphics[width=\textwidth]{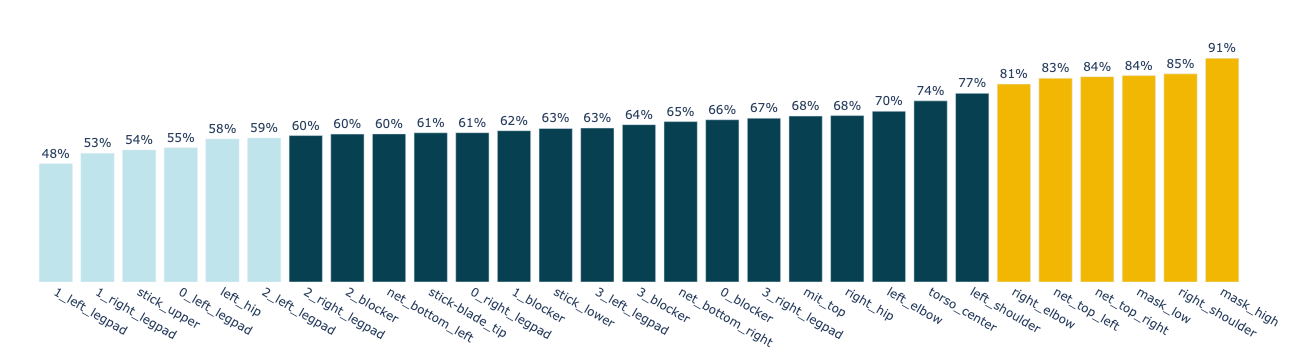}
\caption{Proportion of frames featuring specific keypoints}
\label{fig:dist}
\end{figure}

One challenge posed by this dataset is that not all of the frames contain all of the keypoints. Occurrences of occlusion or missed annotations lead to the absence of certain keypoints in some frames. As a result, the percentage of frames containing specific keypoints varies significantly across different keypoints. This variability is visually illustrated in Fig.~\ref{fig:dist}.

\subsection{GoalieNet Network Design and Training}
The proposed GoalieNet network used to perform joint goalie, equipment, and net pose estimation is a multi-stage network inspired by \cite{li2019rethinking}, where the cross-stage aggregation is leveraged between adjacent stages, and different kernel sizes are used in various levels to improve localization accuracy. We train GoalieNet on all 29 keypoints across the goalie, equipment, and net in a joint fashion, as in practice, the goalie, the equipment, and the net have a strong spatial relationship with each other. Hence, training the network with keypoints of all three entities together allows for more accurate keypoint identification than if trained with separate entities, but this requires the use of an extended bounding box to encapsulate all of these objects. The output of GoalieNet is a total of 29 heatmaps, one heatmap for each keypoint on the goalie, equipment, or net. The predicted coordinates of a given keypoint are the point that is on the 25\% of the length of a vector that connects the heatmap's first and second maximum values.

\subsection{Evaluation metric}
Due to the presence of non-human joints in the goalie pose estimation problem, which differs from standard human pose estimate tasks, we could not apply prevalence indicators like OKS~\cite{lin2014microsoft}. Instead, we utilize the Euclidean distance between the anticipated and actual position divided by the bounding box diameter. We consider the keypoint detected if this distance is less than 0.05.

\section{Results}
\label{sec:res}
\begin{figure}[htbp]
    \centering
    \begin{minipage}[b]{0.45\textwidth}
        \centering
        \includegraphics[width=\textwidth]{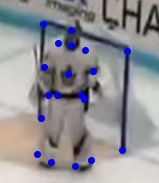}
        %\caption{Figure 1}
        %\label{fig:figure1}
    \end{minipage}
    \hfill
    \begin{minipage}[b]{0.45\textwidth}
        \centering
        \includegraphics[width=\textwidth]{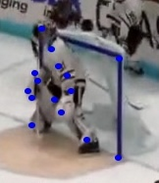}
        %\caption{Figure 2}
        %\label{fig:figure2}
    \end{minipage}
    \caption{Estimated keypoints by GoalieNet}
    \label{fig:res}
\end{figure}

\begin{figure}[htbp]
\centering
\includegraphics[width=\textwidth]{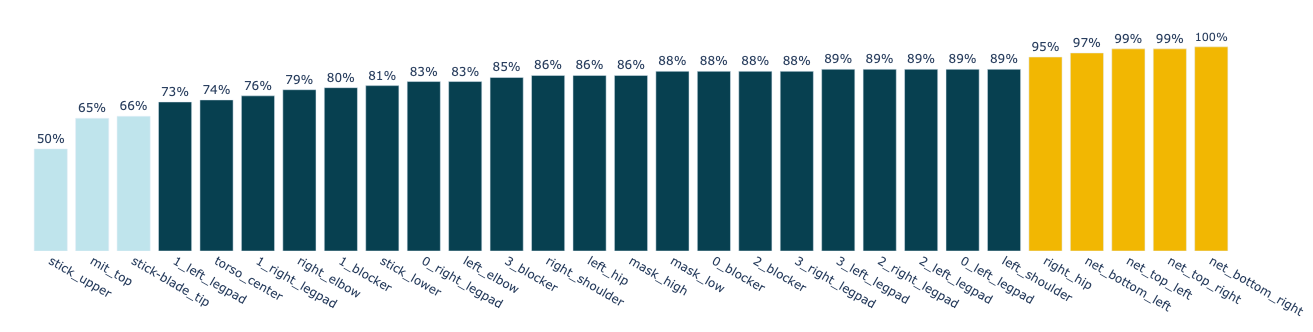}
\caption{This bar plot illustrates the detection accuracy per keypoint. The light blue bars represent the lowest accuracies, while the yellow bars indicate the highest accuracies.}
\label{fig:det-bar}
\end{figure}

\begin{figure}[htbp]
\centering
\includegraphics[width=\textwidth]{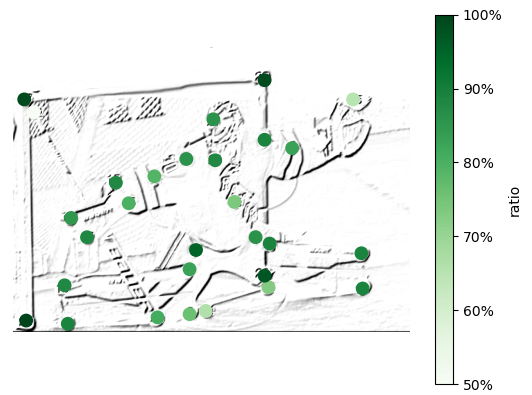}
\caption{The figure illustrates the detection accuracy of each keypoint based on its placement on the goalie's body or equipment.}
\label{fig:det-body}
\end{figure}

Figure~\ref{fig:res} showcases two sample frames with correctly detected keypoints. The detection accuracy across all keypoints is visually represented in Figure~\ref{fig:det-bar}. Additionally, Figure~\ref{fig:det-body} provides insights into the detection accuracy based on the keypoints' positions on the goalie's body or equipment.

The mean accuracy across all keypoints is 84\%. However, there is a wide range of accuracies observed. For instance, the upper part of the stick exhibits a low accuracy of 50\%, whereas the net, with relatively motionless keypoints, and right hip achieve over 95\% accuracy. Notably, the narrow shape and rapid movement of the stick, since it is kept in goalie's hand, pose significant challenges for the network. Nevertheless, the majority of keypoints demonstrate a promising accuracy ranging from 73\% to 89\%.

\section{Discussion and Future Work}
\label{sec:fut}
As seen in \ref{sec:res}, the accuracy of each keypoint varies significantly, with some keypoints being more readily identifiable than others. An important avenue for future research is to investigate how estimating particular keypoint groups might be improved.

The extended bounding box notion adds noise to each image because a noticeable portion of this larger bounding box lacks helpful information. Thus, this problem may be viewed as a joint multi-object pose estimation task where we attempt to separate the goalkeeper, their stick, and the net but still jointly estimate the associated keypoints while at the same time taking into account specific constraints in their joint spatial relationships, such as the fact that the goalie is most of the time in front of the net and holds the stick in their hand. Given that the stick and net are rigid bodies with fixed dimensions, this may be quite effective. Finally, the impact of fixed kernel size should be investigated, since, for example, keypoints on the leg pads are too close together and may require smaller kernels for finer localization.

{\small
\bibliographystyle{ieee_fullname}
\bibliography{main}
}

\end{document}